\documentclass[10pt,twocolumn,letterpaper]{article}

\usepackage{iccv}
\usepackage{times}
\usepackage{epsfig}
\usepackage{graphicx}
\usepackage{amsmath}
\usepackage{amssymb}
\usepackage{comment}
\usepackage{multirow}
\usepackage{dsfont}
\usepackage{caption}
\usepackage{stfloats}
\usepackage{subcaption}

\usepackage[pagebackref=true,breaklinks=true,letterpaper=true,colorlinks,bookmarks=false]{hyperref}

\iccvfinalcopy

\DeclareMathOperator*{\argmin}{arg\,min}

\newcommand{\OURS}{HyperDiffusion\xspace}
\newcommand{\MLP}{MLP\xspace}
\newcommand{\Indicator}{\mathds{1}}

\ificcvfinal\fi

\begin{document}

\title{\OURS: Generating Implicit Neural Fields with Weight-Space Diffusion}

\author{Ziya Erko\c{c}\textsuperscript{1}
\and
Fangchang Ma\textsuperscript{2}
\and
Qi Shan\textsuperscript{2}
\and
Matthias Nie{\ss}ner\textsuperscript{1}
\and
Angela Dai\textsuperscript{1}
\and
\textsuperscript{1}Technical University of Munich \, \textsuperscript{2}Apple\\ \\
\url{https://ziyaerkoc.com/hyperdiffusion}
}

\twocolumn[{%
	\renewcommand\twocolumn[1][]{#1}%
	\maketitle
	\begin{center}
		\includegraphics[width=\linewidth]{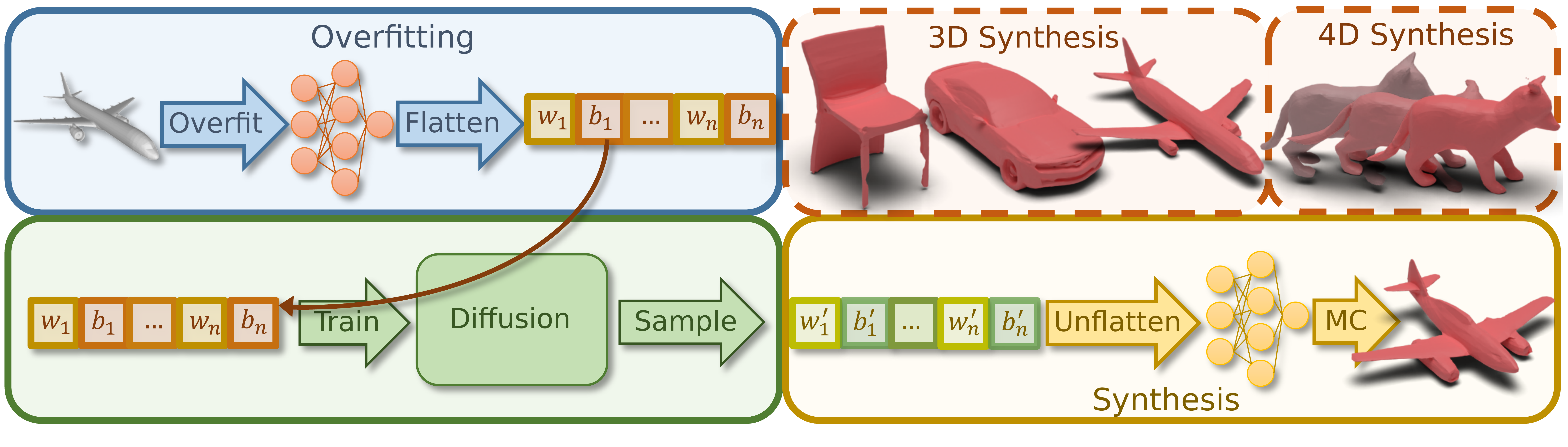}
		\captionof{figure}{
		\OURS{} enables a new paradigm in directly generating neural implicit fields by predicting their weight parameters. 
		We leverage implicit neural fields to optimize a set of MLPs that faithfully represent individual dataset instances (``Overfitting," top-left).
        Our network, based on a transformer architecture, then models a diffusion process directly on the optimized MLP weights (``Diffusion," bottom-left). 
        This enables synthesis of new implicit fields (``Synthesis," bottom-right).
		}
		\label{fig:teaser}
	\end{center}    
}]

\maketitle
\ificcvfinal\thispagestyle{empty}\fi

\begin{abstract}
Implicit neural fields, typically encoded by a multilayer perceptron (MLP) that maps from coordinates (e.g., xyz) to signals (e.g., signed distances), have shown remarkable promise as a high-fidelity and compact representation. 
However, the lack of a regular and explicit grid structure also makes it challenging to apply generative modeling directly on implicit neural fields in order to synthesize new data. To this end, we propose \OURS, a novel approach for unconditional generative modeling of implicit neural fields. \OURS operates directly on MLP weights and generates new neural implicit fields encoded by synthesized MLP parameters. Specifically, a collection of MLPs is first optimized to faithfully represent individual data samples. Subsequently, a diffusion process is trained in this MLP weight space to model the underlying distribution of neural implicit fields.
\OURS enables diffusion modeling over a implicit, compact, and yet high-fidelity representation of complex signals across 3D shapes and 4D mesh animations within one single unified framework.
\end{abstract}

\section{Introduction}
\label{sec:intro}

Recent years have seen profound development in implicit neural field models, demonstrating powerful representations, particularly 3D shape geometry~\cite{park2019deepsdf, chabra2020deepls}, neural radiance fields (NeRF)~\cite{mildenhall2021nerf}, and complex signals with higher-order derivative constraints~\cite{sitzmann2020implicit}. Typically, an implicit neural field\footnote{Implicit neural fields are also referred to as coordinate fields or coordinate-based networks. We use these terms interchangeably.}
maps an input coordinate location in $n$-dimensional space to the target signal domain. For example,
an implicit surface representation
\begin{equation*}
    \{\mathbf{x}\in \mathbb{R}^n | f(\mathbf{x},\theta) = 0\},
\end{equation*}
where $f:\mathbb{R}^n\rightarrow\mathbb{R}$ is typically characterized by a multilayer perceptron (MLP).
Notably, such neural fields efficiently represent sparse high-dimensional data in a relatively low-dimensional MLP weight space.
This continuous mapping enables sampling at arbitrary-resolution for surface representations, eliminating explicit resolution constraints inherent to classical point, mesh, or voxel representations.
One can then easily reconstruct the mesh underlying this compact representation through methods such as Marching Cubes~\cite{lorensen1987marching}.
It has enabled faithful 3D reconstruction where an MLP is optimized to fit a set of point observations~\cite{park2019deepsdf}, as well as preliminary results in higher-dimensional 4D reconstruction and tracking~\cite{li20214dcomplete}.

\begin{figure}[hbtp]
    \centering
    \includegraphics[width=\linewidth]{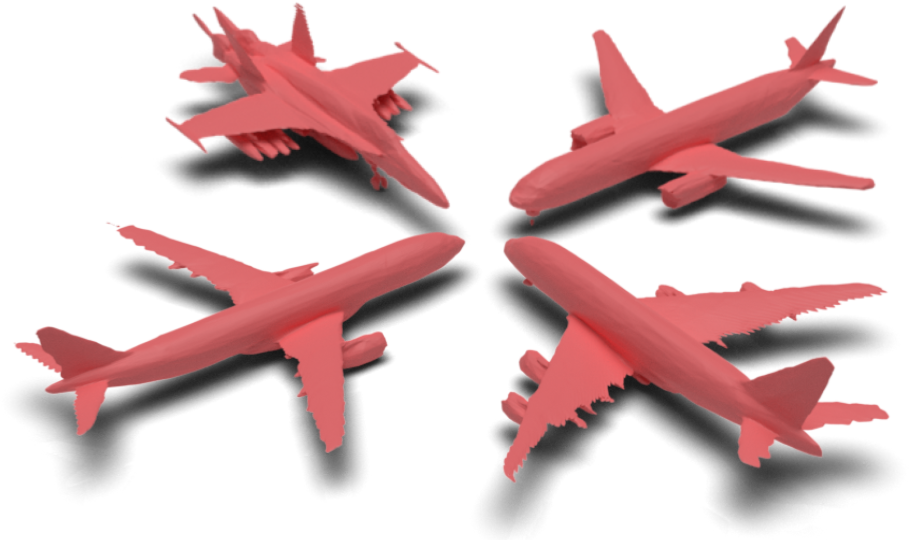}
    (a) Generated 3D shapes.
    
    \includegraphics[width=\linewidth]{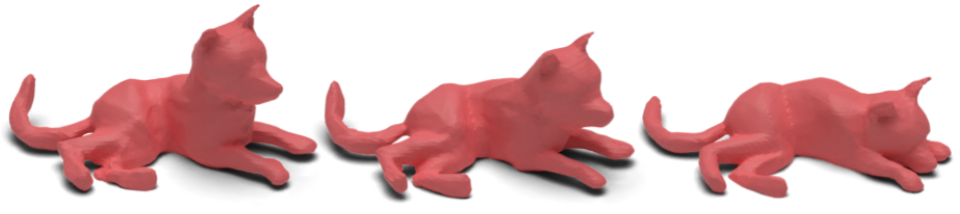}

    (b) Generated 4D animation sequence.
    
    \caption{\OURS is a dimension-agnostic generative model. The same approach can be trained on data of various dimensionalities to synthesize high-fidelity examples. For instance, 
   (a) 3D shapes, and (b) 4D animation sequences (3D and time).}
    \label{fig:multi_planes}
\end{figure}

The aforementioned implicit neural field representation also poses a new opportunity for generative modeling -
that is, rather than model the raw 3D or 4D surface information, we aim to directly model the space of optimized neural field MLPs without relying on any ties to alternative explicit representations (e.g., points, meshes, and voxels) or approximation to latent manifolds (which typically require a pair of  encoder and decoder networks pretrained on large-scale datasets).
In particular, we observe that optimized neural field MLPs typically maintain an extremely compact representation of a high-dimensional 3D or 4D surface.

We thus propose \OURS{}, a new paradigm for generative modeling of neural fields. 
\OURS{} leverages diffusion modeling directly on the weight space of optimized neural fields, enabling generative modeling of high-dimensional data characterized by neural fields.
Our \OURS{} approach is dimension-agnostic, and we apply the same method towards both unconditional 3D and 4D (3D and time) surface generation to demonstrate the power of directly generating neural fields as MLP weights.

Specifically, to model neural fields characterizing $n$-dimensional surface data, we consider a set of MLPs that have been optimized to represent individual instances from a dataset.
We then employ a transformer-based network to model the diffusion process directly on the optimized MLP weights.
This enables generative modeling on a low-dimensional space, and we demonstrate its high-fidelity and diverse generative modeling capabilities, achieving state-of-the-art 3D and 4D surface synthesis.

\medskip
Our contributions can be summarized as follows:
\begin{itemize}
    \vspace{-0.15cm}
    \item We present the first approach to model the space of neural field MLP weights by diffusion modeling, enabling a new paradigm for high-dimensional generative modeling.
    \vspace{-0.15cm}
    \item Our MLP optimization of surface occupancy provides a low-dimensional weight space for  effective unconditional diffusion modeling with a transformer-based architecture for both 3D and 4D surfaces at high fidelity. 
\end{itemize}

\begin{figure*}[t]
    \centering
    \includegraphics[width=\linewidth]{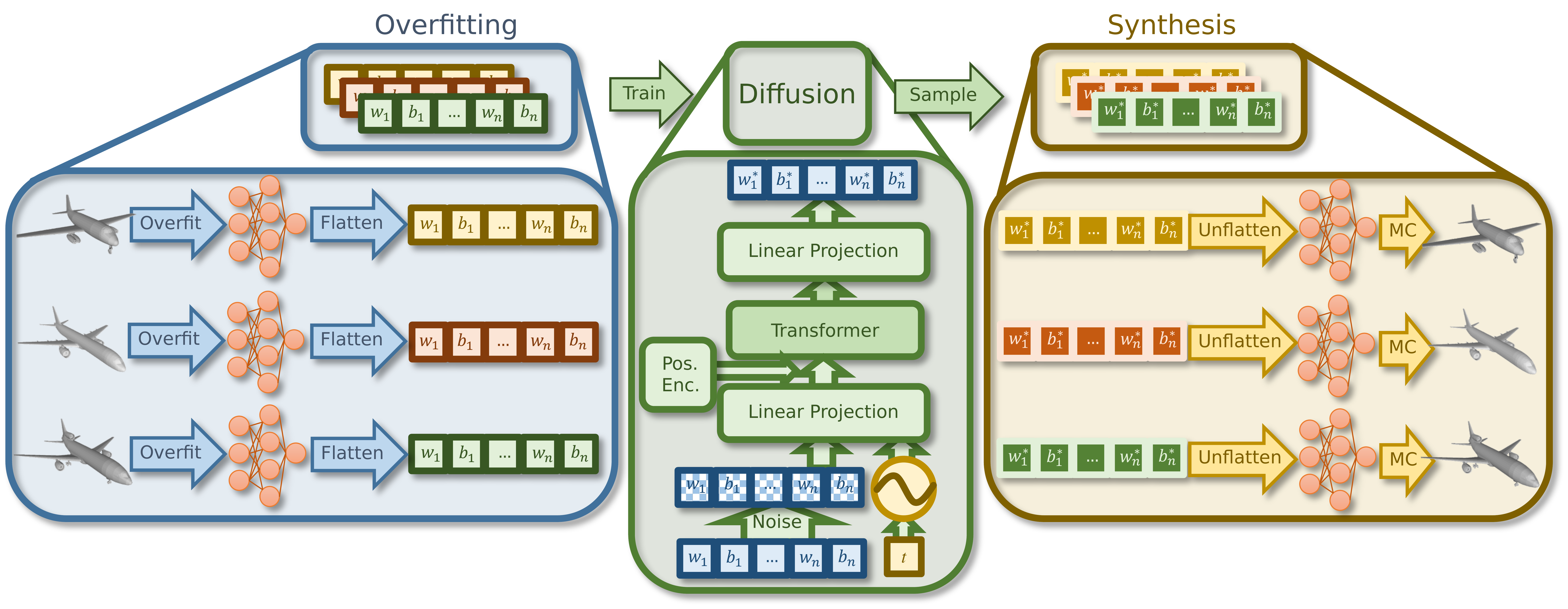}
    \caption{
    Overview of \OURS{}.
    We first fit a set of neural field MLPs to a dataset of 3D or 4D shapes in an overfitting process, producing high-fidelity shape representations that support arbitrary resolutions (left).
    To support the following diffusion process on the MLP weights, we use an arbitrary optimized MLP to initialize the rest of the MLP optimization.
    We then use a transformer-based architecture to model a diffusion process directly on these optimized MLP weights, predicting the denoised MLP weights and biases as flattened vectors (middle).
    This enables synthesis of new neural field representations as their MLP weights, from which meshes can be extracted through Marching Cubes.
    Our approach is agnostic to the resolution of the dataset instances, and we demonstrate effective 3D and 4D shape modeling.
    }
    \label{fig:overview}
\end{figure*}
\section{Related Work}
\label{sec:related}

\paragraph{Neural Implicit Fields}
Neural implicit fields have shown promising results in representing high-fidelity geometry and appearance in 3D. One seminal early work is DeepSDF~\cite{park2019deepsdf}, which encodes the shapes of a class of objects as signed distance functions using a multi-layer fully-connected neural network. Neural Radiance Fields (NeRF)~\cite{mildenhall2021nerf}  encode a coordinate-based radiance field with a \MLP and is capable to producing photo-realistic 2D renderings at novel views through volumetric rendering. To address the bias towards learning low-frequency details in a standard MLP, Fourier features~\cite{tancik2020fourier} and periodic activation functions~\cite{sitzmann2020implicit} have been proposed to improve representation of complex signals.
In this work, we train on the weights of these multi-layer fully-connected neural networks and generate new weights that represent valid signals.

\paragraph{Generative Adversarial Networks (GANs)}
Generative Adversarial Networks (GANs)~\cite{goodfellow2020generative} has been shown to be capable of generating high-resolution 2D images, most notably StyleGAN~\cite{karras2019style, Karras2019stylegan2, Karras2021, Sauer2021ARXIV}. More recently, GANs have been adopted for 3D generation with various underlying representations. For instance, pi-GAN~\cite{chan2021pi} modulates the \MLP network which encodes a radiance field,  EG3D~\cite{chan2022efficient} proposes an efficient tri-plane representation, and GMPI~\cite{zhao2022generative} directly generates multiplane images for efficient training and inference. A similar line of work focuses on generating textures for meshes~\cite{siddiqui2022texturify, gao2022get3d}. A general limitation with GANs is volatile training stability due to the game theoretical nature of generator and discriminator networks competing against each other. This motivates researcher to look into alternative generative models, including Diffusion Process. In this work, we adopt a diffusion model for the MLP weight generation.

\paragraph{2D Diffusion Models}
Diffusion probabilistic models~\cite{ho2020denoising, song2020denoising} have emerged as a powerful alternative to its precursors, such as GANs~\cite{goodfellow2020generative} and energy-based models (EBMs)~\cite{du2019implicit}, for generative modeling. They have been shown to outperform GANs on image synthesis tasks in terms of not only superior quality and fidelity~\cite{dhariwal2021diffusion,rombach2022high}, but also the capability to enable text-conditioning~\cite{nichol2021glide, saharia2022photorealistic}. Specifically, Latent Diffusion~\cite{rombach2021highresolution,https://doi.org/10.48550/arxiv.2204.11824} enables high-resolution image synthesis by applying diffusion in the latent space of pretrained autoencoders. In contrast to Latent Diffusion, our method operates on the \MLP weight space and generates new neural implicit fields.

\paragraph{3D/4D Diffusion Models}
Diffusion models have also been deployed in more challenging tasks beyond image synthesis, including 2D video generation~\cite{ho2022video}, 3D shape and scene generation~\cite{Chan2021, chan2021pi, chou2022diffusionsdf, bautista2022gaudi, poole2022dreamfusion, muller2022diffrf}, and 4D generation with animation of 3D shapes~\cite{singer2023text4d, tevet2022human}. In particular, MDM is a human motion diffusion model where the input is a sequence of human poses \cite{tevet2022human}. They use a transformer architecture  as the denoising network. It takes in a noisy human pose sequence input and denoises it to generate a new motion animation. Unlike their approach, we can directly generate animated meshes instead of just human poses. 

Diffusion modeling in a latent manifold for neural fields has also been concurrently proposed, including DiffusionSDF~\cite{chou2022diffusionsdf}, 3DShape2VecSet~\cite{zhang20233dshape2vecset}, and LION~\cite{zeng2022lion}. Such latent diffusion models require a high-quality latent manifold to be learned, which can be challenging with limited quantities of 3D and 4D data. In contrast, our approach to model the weights of optimized neural fields operates on inherently high quality shape representations that can be extremely well-fit per instance. 

\paragraph{Cross-modality Diffusion Models}
A number of recent publications have been investigating unification of diffusion models across different data dimensions and modalities, including Functa~\cite{dupont2022data}, GEM~\cite{du2021learning}, and GASP~\cite{dupont2021generative}. Most recently, Diffusion Probabilistic Fields~\cite{zhuangdiffusion} proposes a explicit field representation without a latent field parametrization, and formulates the generative model in a single-stage end-to-end training. Our method can be seen as an alternative solution to uniting methods designed across different data dimensions, since the implicit neural fields are essentially dimension-agnostic.

\section{Method Overview}
\label{sec:overview}
\OURS is an unconditional generative model for implicit neural fields encoded by MLPs. 
We operate directly on MLP weights, enabling generation of new neural implicit fields characterized by synthesized MLP parameters. 
Our training paradigm encompasses a two-phase approach, as shown in Figure~\ref{fig:overview}. 

In the first MLP overfitting step, detailed in Section~\ref{sec:overfit}, we optimize a collection of MLPs such that each MLP represents a faithful neural occupancy field of a data sample (e.g., a 3D shape) from the training set.
This enables highly-accurate shape fitting due to the representation power of the neural fields.
The optimized MLP weights are flattened into 1D vectors and passed to a downstream diffusion process as ground truth signals. 

In the second step, detailed in Section~\ref{sec:diffusion}, the aforementioned optimized MLP weights are passed into a diffusion network for training. This diffusion network is domain-agnostic without any assumptions or prior knowledge on the dimensionality of the underlying signal, since its input is a set of flattened MLP weight vectors. After training is completed, new MLP weights, which correspond to a valid neural implicit field, can be synthesized through the reverse diffusion process on a randomly sampled noise signal. 

For 3D and 4D generation, the underlying meshes can be further extracted and visualized with Marching cubes~\cite{lorensen1987marching}. 
\section{Per-Sample MLP Overfitting}
\label{sec:overfit}
In this first step, we optimize MLPs for each data sample from the training set $\left\{S_i, i=1, \dots, N\right\}$ and save their optimized network weights.
Specifically, for a train sample $S_i$, the surface at iso-level $\tau$ is represented as
\begin{align}
\{ &\mathbf{x}\in\mathbb{R}^n,\quad f(\mathbf{x},\theta_i) = \tau \},
\end{align}
where $f$ is an MLP parametrized by $\theta_i \in \mathbb{R}^h$ (\ie, weights and biases in Fig.~\ref{fig:overview}) and $\mathbf{x}$ represents a spatial location. The goal is to minimize the binary cross entropy loss function
\begin{align}
L = \textrm{BCE}(f(\mathbf{x}, \theta_i), o^\textrm{gt}_i(\mathbf{x})),
\end{align}
where $o^\textrm{gt}_i(\mathbf{x})$ is the ground truth occupancy of $\mathbf{x}$ with respect to $S_i$.
Here, we exploit the representation power of neural fields, which can model high-dimensional surfaces with high accuracy.
As these optimized MLPs serve as ground truth for the following diffusion processes, their ability to model high-fidelity shapes is crucial.

Unlike prior methods, we do not require any auto-encoding networks~\cite{bagautdinov2018modeling, bloesch2018codeslam} nor auto-decoding networks (\eg, DeepSDF~\cite{chou2022diffusionsdf}), which typically share the same network parameters across an entire dataset. In contrast, although we use the same MLP architecture for different samples in the training dataset, one set of MLP weights is optimized specifically for each data sample. In other words, there is no parameter sharing, and this per-sample optimization attains the best fidelity possible for the implicit neural field representation.

\paragraph{MLP Architecture and Training}
Our MLP architecture is a standard multi-layer fully-connected network with ReLU activation functions and an input positional encoding~\cite{mildenhall2021nerf}. 
We use 3 hidden layers with 128 neurons each, finally outputting a scalar occupancy value. 
The same MLP architecture is shared across both 3D shape and 4D animation experiments, where each MLP encodes an occupancy field per-train sample. 
This enables dimension-agnostic paradigm for encoding of various data signals, in our case 3D and 4D shapes, where only the positional encoding is adapted for various-dimensional inputs.

To optimize a set of MLP weights and biases for an input 3D shape, we sample points randomly both inside and outside of the 3D surface. 
We normalize all train instances to $[-0.5, 0.5]^3$, and randomly sample $100k$ points within the space. 
To effectively characterize fine-scale surface detail, we further sample $100k$ points near the surface of the mesh. 
Both sets of points are combined and tested for inside/outside occupancy using generalized winding numbers \cite{barill2018fast}; these occupancies are used to supervise the overfitting process. 
We optimize each MLP with a mini-batch size of $2048$ points, trained with a BCE loss for 800 epochs until convergence, which takes $\approx 6$ minutes per shape.

MLP overfitting to 4D shapes is performed analogously to 3D. 
For each temporal frame, we sample $200k$ points and their occupancies, following the 3D shape sampling. 
The sampling process is repeated for each frame of an animation sequence. 
We optimize one set of MLP weights and biases for each animation sequence to represent each 4D shape.

\paragraph{Weight Initialization}
In order to encourage a smooth diffusion process over the set of optimized MLP weights, we guide the MLP optimization process with consistent weight initialization.
That is, we initially optimize one set of MLP weights and biases $\theta_1$ to represent the first train sample $S_1$, and use the optimized weights of $\theta_1$ to initialize optimization of the rest of the MLPs $\{\theta_2,...,\theta_N\}$.
\section{MLP Weight-Space Diffusion}
\label{sec:diffusion}

We then model the weight space of our optimized MLPs through a diffusion process.
We consider each set of optimized MLP weights and biases $\{\theta_i\}$ as a flattened 1D vector.
We use a transformer architecture $\mathcal{T}$, following \cite{peebles2022learning}, for our denoising network. 
As transformers have been shown to elegantly handle long vectors in the language domain, we find it to be a suitable choice for modeling the MLP weight space.
$\mathcal{T}$ predicts the denoised MLP weights directly, rather than the noise.
Our $h$-dimensional vectors $\{\theta_i\}$ of MLP weights and biases are input to $\mathcal{T}$.
Each $\theta_i$ is divided into 8 tokens by MLP layers, to be encoded by $\mathcal{T}$.

Modeling the MLP weights as a 1D vector for diffusion enables a general formulation for modeling neural fields, as the MLP weights are agnostic to varying-dimensional data.
This makes \OURS{} flexible to a variety of neural field representations; in particular, we observe the neural field ability to compactly represent high-dimensional shape data and demonstrate  generative modeling of MLPs representing 3D and 4D shapes, respectively.

During diffusion modeling, we apply standard gaussian noise $t$ times to each vector $\theta$. 
The noisy vector, along with the sinusoidal embedding of $t$, are then input to a linear projection. The projections are then summed up with a learnable positional encoding vector.
Then, the transformer outputs denoised tokens that we pass through a final output projection to produce the predicted denoised MLP weights $w^*$.
We train with a Mean Squared Error (MSE) loss between the denoised weights $\theta^*$ and the input weights $\theta$.

\begin{figure*}[hbtp]
    \includegraphics[width=\linewidth]{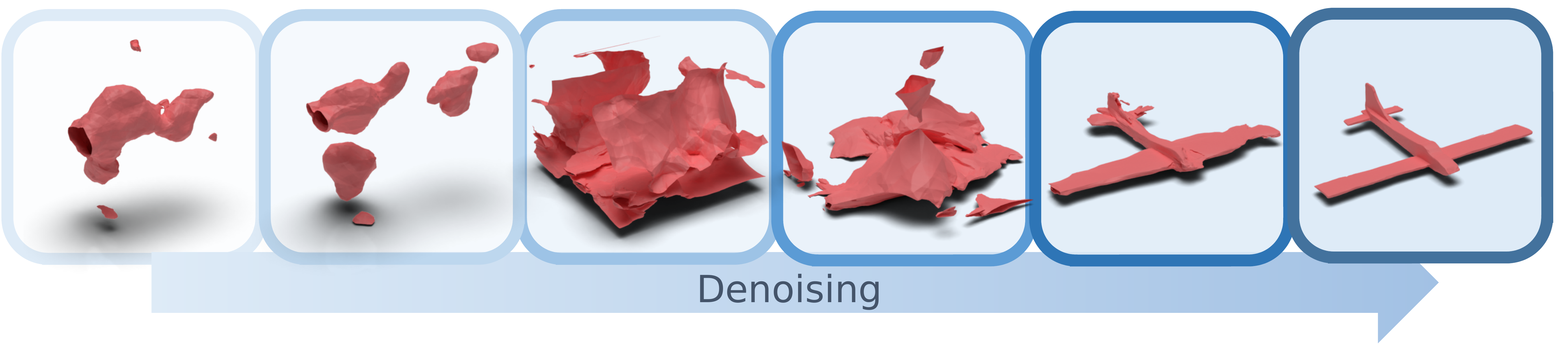}
    \caption{
    Denoising MLP parameters at various time steps, visualized with their corresponding shapes, from random noise (left) to fully denoised (right). The image shows the gradual change from 0 denoising steps, which are generated from random MLP weights, up to 500 steps corresponding to fully-denoised shape. More specifically, we leverage the DDIM \cite{song2020denoising} sampling strategy. Interestingly, noisy MLP weights do not necessarily correspond to a valid 3D shape; however, the iterative application of the denoising operator eventually converges to a quality output shape.  
    }
    \label{fig:denoising_vis}
\end{figure*}

We illustrate the denoising process for MLPs representing 3D shapes in Figure~\ref{fig:denoising_vis}. 
Noisy MLPs correspond to invalid shapes, which are denoised to MLPs that represent valid 3D surfaces. We employ Denoising Diffusion Implicit Models (DDIM) \cite{song2020denoising} to sample new MLPs from the diffusion process.

\paragraph{Implementation Details}
Our 3-layer 128-dim MLPs  contain $\approx 36$k parameters, which are flattened and tokenized for diffusion.
We use an  AdamW~\cite{loshchilov2017decoupled} optimizer with batch size $32$ and initial   learning rate of $2e^{-4}$, which is reduced by $20$\% every $200$ epochs.
We train for $\approx 4000$ epochs until convergence, which takes $\approx 4$ days on a single A6000 GPU.

\section{Results}
\label{sec:results}

\subsection{Datasets}

For 3D shape generation, we use the car, chair, and airplane categories of the ShapeNet~\cite{chang2015shapenet} dataset. The car, chair and airplane categories have 3533, 6778, and 4045 shapes respectively. For the 4D shape generation task, we use 16-frame animal animation sequences from the  DeformingThings4D~\cite{li20214dcomplete} dataset, comprising  1772 sequences. 
For both datasets, we split the data into non-overlapping partitions, including training (80\%), validation (5\%) and testing (15\%) subsets.

\subsection{Voxel-based Diffusion Baseline}
\label{sec:vox-baseline}
While several existing methods tackle 3D shape generation, unconditional 4D shape generation remains underexplored.
Thus, in addition to existing 3D baselines, we introduce a voxel-based diffusion model as a baseline for both 3D and 4D shape generation.

3D shapes are represented as dense occupancy grids, and a 3D UNet denoising network is applied on the 3D voxel grids.
4D shapes are represented similarly, with each frame of an animation sequence voxelized to a 3D occupancy grid, producing a 4D occupancy grid representing the full sequence.
We use the same 3D UNet as a denoising network to synthesize 4D animation, as a 4D UNet became computationally intractable.

For our experiments, we use a voxel resolution of $24^3$ for 3D shapes and $16\times 24^3$ for 4D shapes (the maximum spatial resolution such that 4D grids could be tractably trained).

\subsection{Evaluation Metrics}

Evaluation of unconditional synthesis of 3D and 4D shapes can be challenging due to lack of direct correspondence to ground truth data.
We thus follow prior works \cite{zeng2022lion, luo2021diffusion, zhou20213d} in evaluating  Minimum Matching Distance (MMD), Coverage (COV), and 1-Nearest-Neighbor Accuracy (1-NNA). For MMD, lower is better; for COV, higher is better; for 1-NNA, 50\% is the optimal.
\begin{align*}
\text{MMD}(S_g, S_r) &= \frac{1}{\vert S_r \vert} \sum_{Y \in S_r} \min_{X \in S_g} D(X, Y), \\
\text{COV}(S_g, S_r) &= \frac{\vert \{ \argmin_{Y \in S_r} D(X, Y) \vert X \in S_g \} \vert}{\vert S_r \vert}, \\
\text{1-NNA}(S_g, S_r) &= \frac{\sum_{X \in S_g} \Indicator[N_X \in S_g] + \sum_{Y \in S_r} \Indicator[N_Y \in S_r] }{\vert S_g \vert + \vert S_r \vert},
\end{align*}
where in the 1-NNA metric $N_X$ is a point cloud that is closest to $X$ in both generated and reference dataset, i.e., 
$$N_X = \argmin_{K \in S_r \cup S_g} D(X, K)$$

We use a Chamfer Distance (CD) distance measure $D(X,Y)$ for computing these metrics in 3D, and report CD values  multiplied by a constant $10^2$. 
To evaluate 4D shapes, we extend to the temporal dimension for $T$ frames:
$$D(X, Y) = \frac{1}{T}\sum_{t=0}^{T - 1} CD(X[t], Y[t]).$$
We note that the MMD metric has been discussed to be unreliable as a measure of generation quality~\cite{zeng2022lion}, and thus also consider perceptual metric for 3D shape generation.
In particular, we follow \cite{zhang20233dshape2vecset} and compute a Frechet Pointnet++ Distance~\cite{qi2017pointnet++} (FPD), analogous to the commonly used Frechet Inception Distance (FID score)~\cite{heusel2017gans} in the image domain. 
FPD instead uses a  3D Pointnet++ network (trained on the ModelNet~\cite{wu20153d}) for feature extraction from generated shapes. For FPD, lower scores are better.

For evaluation, each synthesized and ground truth shape is normalized by its mean and standard deviation on a per-shape basis.
To evaluate point-based measures, we sample 2048 points randomly from all baseline outputs; for our approach and the voxel baseline, points are sampled from the extracted mesh surface, and for point cloud baselines, points are sampled directly from the synthesized outputs. 

\begin{figure}[htbp]
    \centering
    \includegraphics[width=0.95\linewidth]{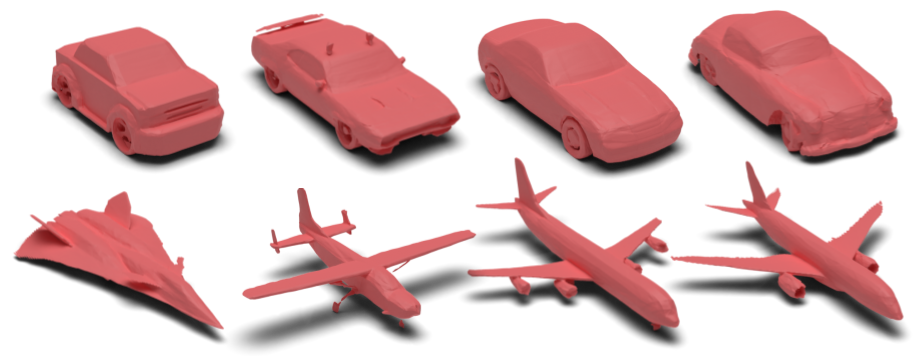}
    \caption{Synthesized 3D shapes with \OURS trained on ShapeNet~\cite{chang2015shapenet}. 
    Each shape is represented with a 3-layer 128-dim MLP. We extract the underlying isosurface from the MLPs with Marching Cubes \cite{lorensen1987marching}.
    }
    \label{fig:generation_3d}
\end{figure}

\subsection{Unconditional 3D Generation}

We demonstrate \OURS{} on unconditional neural field generation of 3D shapes. For our method, we show output meshes extracted with Marching Cubes~\cite{lorensen1987marching}.

\begin{figure}[tp]
    \centering
    \includegraphics[width=1.05\linewidth,trim={0.7cm 0.5cm 0 0cm}]{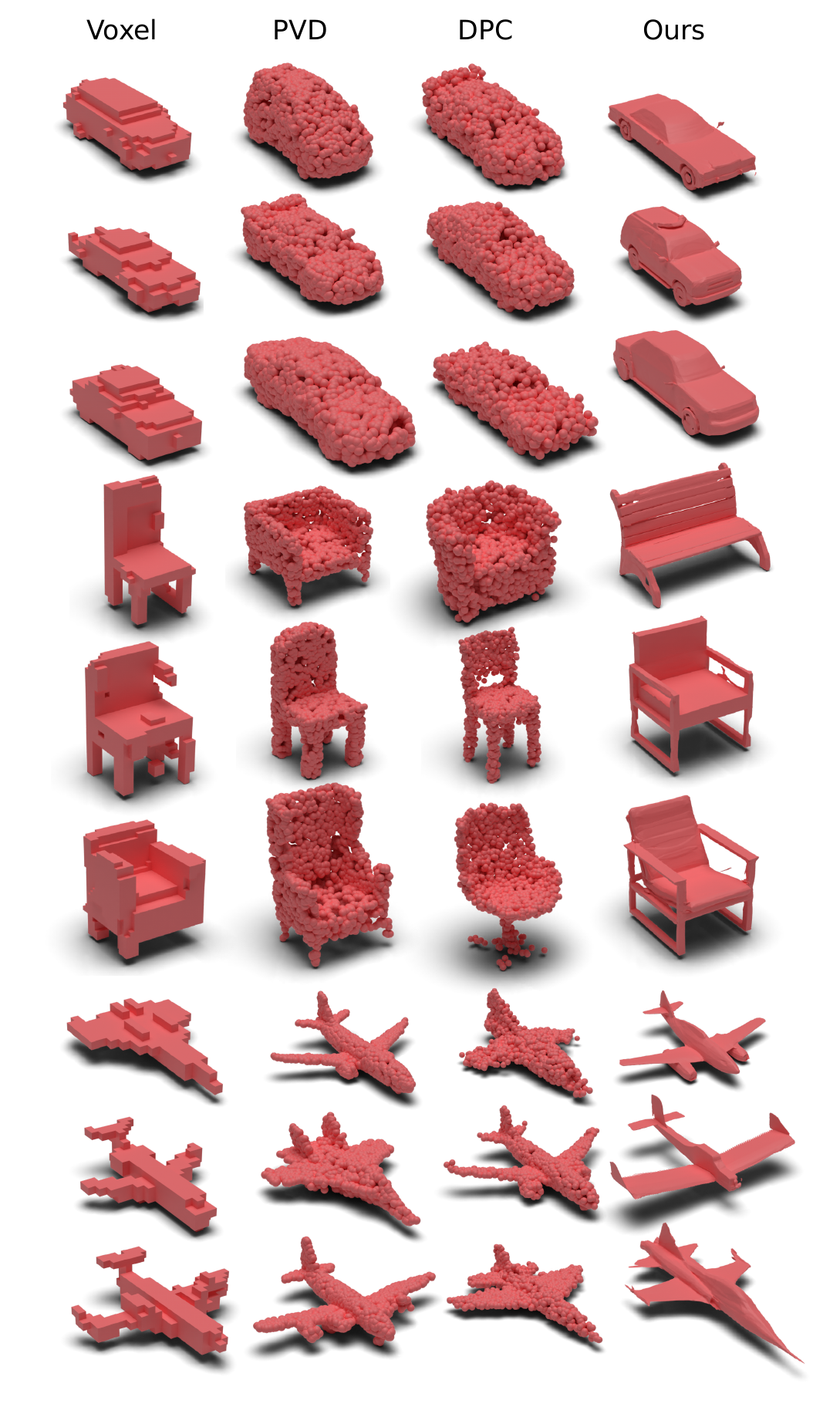}
    \caption{
    Qualitative comparison for 3D shape generation.
    Voxel-based diffusion produces relatively low-resolution outputs, while state-of-the-art PVD~\cite{zhou20213d} and DPC~\cite{luo2021diffusion} synthesize discrete point clouds. 
    In contrast, our neural field synthesis can generate a high-quality, continuous surface representation easily extracted as a mesh.
    }
    \label{fig:3d_comparison}
\end{figure}

\paragraph{Comparison to state of the art}
We compare with state-of-the-art 3D shape generation methods Point-Voxel Diffusion~\cite{zhou20213d} (PVD), Diffusion Point Cloud~\cite{luo2021diffusion} (DPC), as well as our voxel baseline in Table~\ref{tab:quant_3d_uncond_gen}.
\OURS{} achieves improved performance over these baselines in all shape categories and all evaluation metrics except MMD, which has been noted to be less reliable due to lack of sensitivity to low-quality results~\cite{zeng2022lion}.
Notably, we achieve significantly improved FPD, which captures the perceptual quality of our synthesized shapes.
Figure~\ref{fig:3d_comparison} further shows our capability to represent high-resolution detail as captured in our synthesized neural fields, in comparison with baselines.

\paragraph{What is the effect of MLP overfitting weight initialization?}
We analyze the effect of our weight initialization strategy for MLP optimization in Table~\ref{tab:ablation_init}, in comparison with random initialization per-MLP. 
Instead of initializing each MLP completely independently, our consistent weight initialization from a single optimized MLP produces improved results, particularly for the perceptual FPD metric. This could be attributed to the fact that with a consistent initialization the MLP weights stay relatively close to each other.

\paragraph{What is the impact of MLP positional encoding?}
We ablate the effect of positional encoding in our per-instance MLP representations in Table~\ref{tab:ablation_pe}, where No PE denotes the same MLP architecture without any positional encoding.
We see that applying positional encoding to the input coordinates significantly improves generation quality.

\begin{figure}[htb!]
    \centering
    \vspace{0.3cm}
    \includegraphics[width=\linewidth]{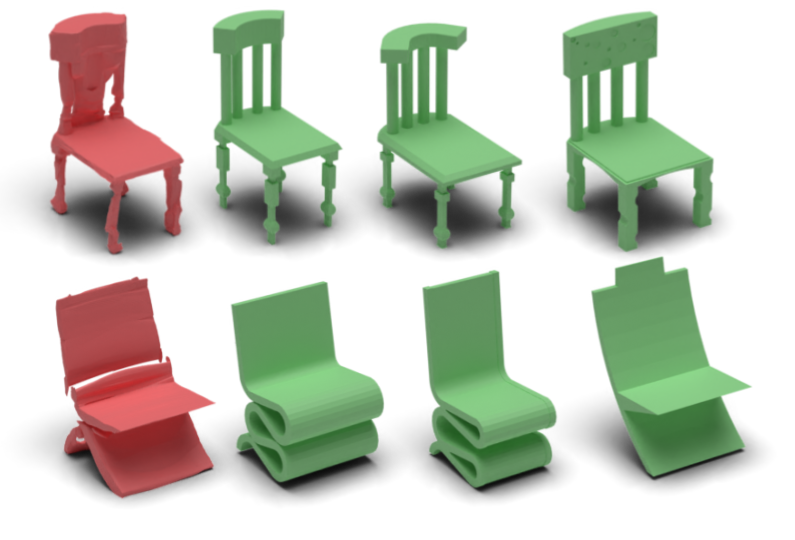}
    \caption{Novel shape generation vs nearest neighbor retrieval. For generated shapes (red) from our method, we look up the top-3 nearest neighbors (green) from the training set based on the Chamfer distance. As shown, our method does not simply memorize train samples and can generalize to novel shapes.}
    \label{fig:novel}
\end{figure}

\paragraph{Novel shape synthesis.}
We explore the degree of novelty in our generated neural fields in 3D shape generation.
Figure~\ref{fig:novel} our synthesized shapes (red) in comparison to the top-3 nearest neighbours by Chamfer distance from the training set (green). 
We show that our synthesized neural fields represent shapes  not already present in the training set, but rather novel shape data.
This is critical to the core idea of our approach since it supports the claim that running denoising duffusion over neural network parameter weights facilitates generalizablity within the underlying data distribution of 3D meshes.

\begin{table*}[]
\centering
\begin{tabular}{cccccc}
\hline
Category & Method & MMD $\downarrow$ & COV (\%) $\uparrow$ & 1-NNA (\%) $\downarrow$ & FPD $\downarrow$ \\ \hline
\multirow{4}{*}{Airplane} & Voxel Baseline & \ \ 6.0 & 28 & 94.1 & 38.9 \\
 & PVD~\cite{zhou20213d} & \ \ 3.4 & 39 & 76.3 & \ \ 5.8 \\
 & DPC~\cite{luo2021diffusion} & \ \ \textbf{3.1} & 46 & 74.7 & 18.7 \\
 & Ours & \ \ 3.4 & \textbf{49} & \textbf{69.3} & \ \ \textbf{3.5} \\ \hline
\multirow{4}{*}{Car} & Voxel Baseline & \ \ 4.9 & 13 & 98.7 & 10.4 \\
 & PVD~\cite{zhou20213d} & \ \ 3.5 & 27 & 76.8 & \ \ 4.5  \\
 & DPC~\cite{luo2021diffusion} & \ \ \textbf{3.3} & 33 & 82.4 & \ \ 7.6 \\
 & Ours & \ \ 3.4 & \textbf{36} & \textbf{73.1} & \ \ \textbf{2.6} \\ \hline
\multirow{4}{*}{Chair} & Voxel Baseline & 11.8 & 28 & 80.6 & 30.6 \\
 & PVD~\cite{zhou20213d} & \ \ 6.8 & 42 & 58.3 & \ \ 3.5 \\
 & DPC~\cite{luo2021diffusion} & \ \  \textbf{6.3} & 44 & 61.4 & 26.0 \\
& Ours & \ \ 7.1 & \textbf{53} & \textbf{54.1} & \ \ \textbf{1.7} \\ \hline
\end{tabular}
\caption{
Quantitative comparison on unconditional 3D shape generation for the airplane, car and chair categories  from ShapeNet~\cite{chang2015shapenet}. 
Our synthesized neural fields outperform the baseline volumetric method and prior state of the art~\cite{zhou20213d,luo2021diffusion}, particularly on the perceptual FPD metric most representative of visual quality.
}
\label{tab:quant_3d_uncond_gen}
\end{table*}

\begin{table}[]
    \centering
\resizebox{0.5\textwidth}{!}{
    \begin{tabular}{cccc}
    \hline
    Methods & MMD $\downarrow$ & COV (\%) $\uparrow$ & 1-NNA (\%) $\downarrow$ \\ \hline
    Voxel Baseline & 21.9 & 35 & 85 \\ 
    Ours & \textbf{15.5}  & \textbf{45} & \textbf{62} \\ \hline
    \end{tabular}
    }
    \caption{
    Quantitative evaluation of 4D unconditional generation of animation sequences from temporally deforming 3D shapes. 
    \OURS{} enables a compact, high-fidelity representation space for synthesis, producing much more detailed, high-quality results than voxel-based diffusion.
    }
    \label{tab:quant_4d_uncond_gen}
\end{table}

\begin{table}[]
    \centering
\resizebox{0.5\textwidth}{!}{
    \begin{tabular}{c c c c c}
    \hline
    Init. & MMD $\downarrow$ & COV (\%) $\uparrow$ & 1-NNA (\%) $\downarrow$ & FPD $\downarrow$ \\ \hline
    Random & \textbf{3.34} & 49 & 70.4 & 4.01 \\
    1st MLP & 3.45 & \textbf{49} & \textbf{69.3} & \textbf{3.49} \\ \hline
    \end{tabular}
    }
    \caption{Ablation on weight initialization for 3D shape generation (airplanes). 
    Using a consistent initialization from a single optimized MLP leads to moderately improved results, most noticeably in the perceptual FPD metric.
    }
    \label{tab:ablation_init}
\end{table}

\begin{table}[]
    \centering
\resizebox{0.5\textwidth}{!}{
    \begin{tabular}{ccccccc}
    \hline
     & MMD $\downarrow$ & COV (\%) $\uparrow$ & 1-NNA (\%) $\downarrow$ & FPD $\downarrow$ \\ \hline
    No PE & 3.75 & 44 & 76.5 & 6.40 \\
    With PE & \textbf{ 3.45} & \textbf{49} & \textbf{69.3} & \textbf{3.49} \\ \hline
    \end{tabular}
    }
    \caption{Ablation on positional encoding in the MLP for 3D shape generation (airplanes). Without positional encoding (No PE), performance noticeably degrades.
    }
    \label{tab:ablation_pe}
\end{table}

\begin{figure}
    \newcommand\x{0.49}
    \centering
    \vspace{0.3cm}
    \includegraphics[width=0.86\linewidth]{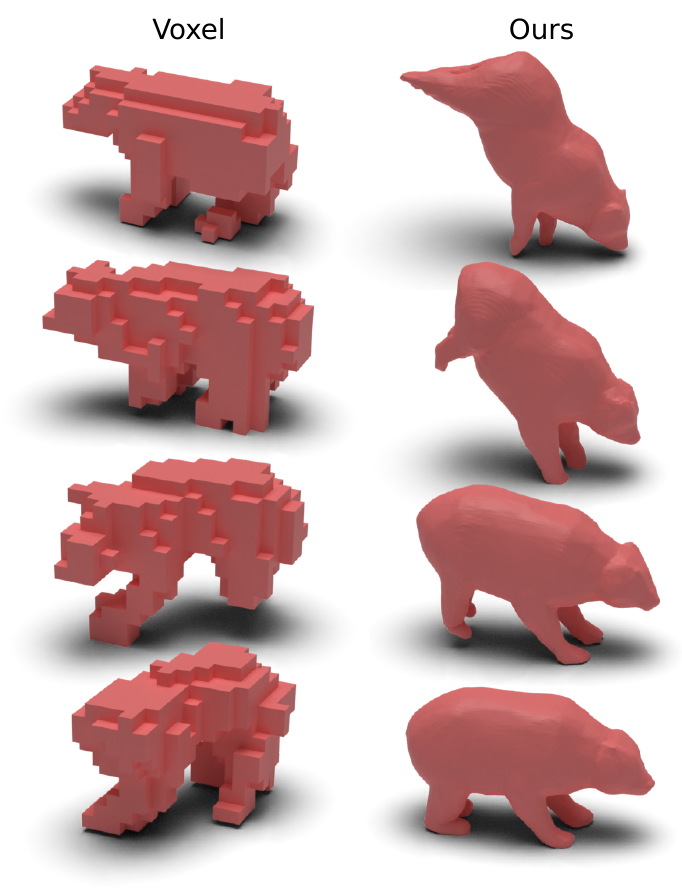}
    \caption{
    Qualitative comparison of 4D animation synthesis. 
    Our neural field synthesis generates not only more detailed animations but also achieves smoother temporal consistency. We also refer to our video for the animated shape results.
    }
    \label{fig:gen_animations}
\end{figure}

\subsection{Unconditional 4D Generation}

As \OURS{} can model neural fields that represent arbitrary dimension data, we exploit the compact representation space of neural fields to model 4D sequences of deforming 3D shapes. The fourth dimension here corresponds to time. 
Table~\ref{tab:quant_4d_uncond_gen} and Figure~\ref{fig:gen_animations} show that our approach can generate much higher quality animation sequences with more detailed representations than a voxel-based diffusion, which becomes quickly bound by its quartic growth in dimensionality. In Figure~\ref{fig:gen_animations}, we visualize 4  out of 16 generated frames for both animation sequences. The full animation sequences in our video and website show that we can generate temporally consistent 4D animations corresponding to meaningful actions (\textit{e.g.,} jumping, rotating, and resting). Shape integrity is preserved throughout the generated animation sequence.

\section{Limitations}
While \OURS{} shows a promising approach towards directly generating neural fields, several limitations remain.
For instance, the diffusion process currently operates only on optimized MLP parameters, without knowledge of any surface reconstruction.
Our initial experiments with a naive secondary reconstruction loss did not improve performance, but we believe a more sophisticated formulation to encourage the diffusion process to be aware of the surface being represented would improve its generative modeling capabilities.
Additionally, we operate on datasets of individual MLPs, while neural implicit scene representations for large-scale environments \cite{jiang2020local,peng2020convolutional} typically employ MLPs on a grid for greater spatial capacity.
Extending to modeling multiple MLPs could enable larger-scale scene surface modeling.
\section{Conclusion}
\label{sec:conclusion}
We proposed \OURS, a new generative modeling paradigm for neural implicit fields. 
We exploit the compact representation power of neural fields for modeling high-dimensional surface data, and  model the weight space of the neural fields with a diffusion process.
This enables high-fidelity surface representations by optimizing neural field MLPs to fit to individual train samples, and using the optimized MLP weights to train the downstream diffusion model. 
We can then synthesize new neural fields as their MLP weights from the diffusion model -- a low-dimensional representation that decodes to high-fidelity shape surfaces for 3D shapes and 4D animation sequences of deforming shapes.
Overall, we believe that our method is a first step to open up new possibilities for generative modeling of high-dimensional, complex data, and alternative representations in the context of diffusion models.

\smallskip
\noindent \textbf{Acknowledgements.}
This work was supported by the Bavarian State Ministry of Science and the Arts coordinated by the Bavarian Research Institute for Digital Transformation (bidt), the ERC Starting Grant Scan2CAD (804724), and the German Research Foundation (DFG) Research Unit “Learning and Simulation in Visual Computing.” Apple was not involved in the evaluations and implementation of the code.

{\small
\bibliographystyle{ieee_fullname}
\bibliography{egbib}
}

\clearpage

\section{Appendix}

\subsection{Additional Qualitative Results}
\begin{figure}
 \centering
 \includegraphics[width=0.98\linewidth]{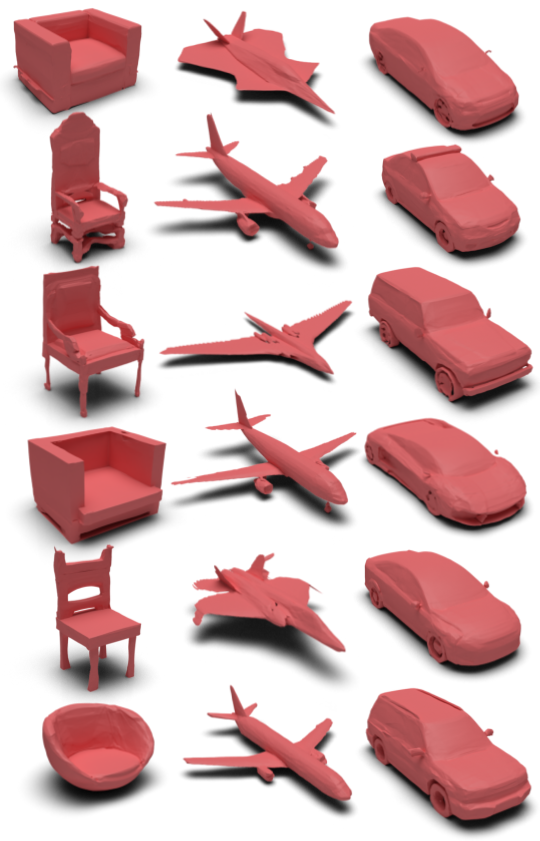}
 \caption{Additional unconditional 3D shape generation results.}
 \label{fig:supplementary_3d}
\end{figure}

\begin{figure}
 \centering
 \includegraphics[width=\linewidth]{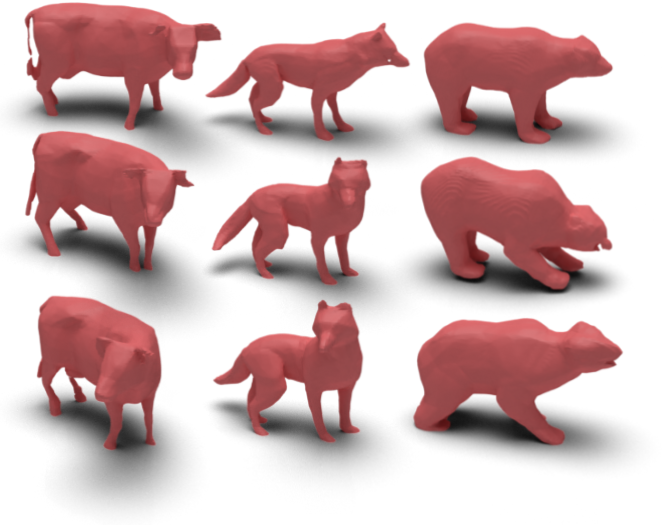}
 \caption{Additional unconditional 4D animation sequence generation results. We refer to our website for animated shape results.}
 \label{fig:supplementary_4d}
\end{figure}

We provide additional unconditional generation results on 3D and 4D generation in Figure~\ref{fig:supplementary_3d} and Figure~\ref{fig:supplementary_4d}. We can generate diverse sets of shapes in both 3D and 4D settings. Resulting meshes are clean, smooth, and can be readily used in any 3D design software and game engines. Although we can output 16 frames for animation sequences, we only show 3 frames in Figure~\ref{fig:supplementary_4d}. Full animation sequences are available in our website and video.

\subsection{Implementation Details}
We use the diffusion and transformer architecture implementations of \cite{peebles2022learning}, which are modified versions of OpenAI and minGPT implementations, respectively. We have a pre-determined MLP structure which consists of 3 hidden layers, each with 256 neurons. To process the MLPs with a transformer architecture, we first flatten the MLP weights into a 1D vector. Additionally, as a way to establish correspondence between components within the 1D vector and the MLP layers (\eg, first \textit{n} values are weights of the first layer), each layer is considered as two tokens, one for its weights and the other for its biases. Hence, in total we have 8 tokens coming from weights and biases. Thanks to this decomposition, transformer may figure out interaction between weights and biases across different layers during training. We also have one additional token representing the sinusoidal embedding of the timestep value. During synthesis of new samples, we again decompose the generated 1D vector into each layer's weights and biases, and load them into the same MLP structure.

Our transformer has 2880 hidden size (i.e., the size of each token after linear projection), 12 layers, and 16 self-attention heads. We use 500 diffusion timesteps in our implementation and a linear noise scheduler ranging between $1e^{-4}$ and $2e^{-2}$. For the sampling strategy, we used DDIM~\cite{song2020denoising} and do not skip any timesteps during sampling. In addition, our denoising network directly predicts the denoised version, following \cite{peebles2022learning}. We observe that $\approx 15\%$ of airplane, $\approx 16\%$ of chair and $\approx 51\%$ of car shapes in our train split of ShapeNet~\cite{chang2015shapenet} contain major self-intersections in their original shape mesh faces, and so we exclude them from the training set for both our approach as well as for all baselines. 

We also apply de-duplication to generated 3D shape results. Note that this has not been applied to baseline approaches, since their quantitative performance degraded with de-duplication. We achieve de-duplication by sampling twice the necessary amount and removing the ones that are very close to each other.
 

\end{document}